\documentclass[conference]{IEEEtran}

\IEEEoverridecommandlockouts
\usepackage{amsmath}
\usepackage{amsfonts}
\usepackage{amssymb}
\usepackage{amsthm}
\usepackage{todonotes}
\usepackage{algorithm}
\usepackage{algpseudocodex}
\usepackage{graphicx}
\usepackage{svg}
\usepackage{tikz}
\usetikzlibrary{positioning}
\usepackage{float}
\usepackage{subfig}
\usepackage{tcolorbox}
\usepackage{bbm}
\usepackage[normalem]{ulem}
\usepackage{bm}
\usepackage{booktabs}
\usepackage{url}
\usepackage{enumitem}

\newtheorem{definition}{Definition}

\newtheorem{remark}{Remark}

\algrenewcommand\algorithmicrequire{\textbf{Input:}}
\algrenewcommand\algorithmicensure{\textbf{Output:}}
\algnewcommand\algorithmicforeach{\textbf{for each}}
\algdef{S}[FOR]{ForEach}[1]{\algorithmicforeach\ #1\ \algorithmicdo}

\DeclareMathOperator*{\argmax}{arg\,max}
\DeclareMathOperator*{\argmin}{arg\,min}

\begin{document}

\title{Optimizing Byzantine Node Placement in Decentralized Federated Learning}
\author{\IEEEauthorblockN{Edoardo Gabrielli}
\IEEEauthorblockA{Dipartimento di Ingegneria Informatica,\\ Automatica e Gestionale\\
Sapienza University of Rome\\
Rome, Italy\\
}
\and
\IEEEauthorblockN{Gabriele Tolomei}
\IEEEauthorblockA{Dipartimento di Informatica\\
Sapienza University of Rome\\
Rome, Italy\\
}
}

\makeatletter\def\@IEEEpubidpullup{6.5\baselineskip}\makeatother

\maketitle

\begin{abstract}
Security evaluations of decentralized federated learning (DFL) typically focus on how Byzantine participants behave, while largely overlooking which participants are compromised. Yet, because aggregation is distributed over a communication graph, the placement of Byzantine nodes determines how malicious influence propagates through the network. We therefore treat Byzantine placement as an explicit adversarial decision and formulate the attacker's objective as selecting, under a fixed compromise budget, the set of participants that maximizes its finite-time impact on honest nodes. To approximate this objective without executing the learning process for every candidate placement, we introduce Byzantine Placement Influence (BPI), a set-level measure derived from the actual gossip dynamics that quantifies the cumulative exposure of honest nodes to Byzantine sources over the training horizon. Unlike placement criteria based on node centrality heuristics, BPI directly accounts for weighted multi-hop propagation and interactions among compromised nodes. We develop efficient algorithms for optimizing BPI and evaluate them across six heterogeneous graph families, untargeted model poisoning, and backdoor attacks. BPI-guided placements consistently identify highly damaging configurations across different network structures and remain effective when the linear gossip assumption is relaxed through Byzantine-robust aggregation. Our results show that Byzantine placement is a critical but under-modeled dimension of DFL threat models and robustness evaluations.
\end{abstract}

\IEEEpeerreviewmaketitle

\newcommand{\edo}[1]{\todo[inline,color=teal!60]{{\bf Edoardo:} #1}}
\newcommand{\gabri}[1]{\todo[inline,color=red!60]{{\bf Gabri:} #1}}
\newcommand{\antho}[1]{\todo[inline,color=orange!60]{{\bf Anthony:} #1}}

\newcommand{\R}{\mathbb{R}}                
\newcommand{\E}{\mathbb{E}}                
\newcommand{\Prob}{\mathbb{P}}             
\newcommand{\calV}{\mathcal{V}}            
\newcommand{\calN}{\mathcal{N}}            
\newcommand{\calH}{\mathcal{H}}            
\newcommand{\calB}{\mathcal{B}}            
\newcommand{\calW}{\mathcal{W}}            
\newcommand{\calO}{\mathcal{O}}            
\newcommand{\calD}{\mathcal{D}}            
\newcommand{\calG}{\mathcal{G}}
\newcommand{\calE}{\mathcal{E}}
\newcommand{\calA}{\mathcal{A}}
\newcommand{\calC}{\mathcal{C}}
\newcommand{\calJ}{\mathcal{J}}
\newcommand{\calL}{\mathcal{L}}
\newcommand{\calS}{\mathcal{S}}

\newcommand{\bA}{\mathbf{A}}               
\newcommand{\bW}{\mathbf{W}}               
\newcommand{\bI}{\mathbf{I}}               
\newcommand{\bone}{\mathbf{1}}             
\newcommand{\btheta}{\boldsymbol{\theta}}  
\newcommand{\bx}{\mathbf{x}}               
\newcommand{\by}{\mathbf{y}}              
\newcommand{\bz}{\mathbf{z}}              
\newcommand{\bg}{\mathbf{g}}              
\newcommand{\bG}{\mathbf{G}}              
\newcommand{\bdelta}{\bm{\delta}}         
\newcommand{\sbdelta}{\bm{\hat{\delta}}}         
\newcommand{\bDelta}{\bm{\Delta}}         
\newcommand{\sbDelta}{\bm{\hat{\Delta}}}         
\newcommand{\bv}{\mathbf{v}}               
\newcommand{\bxi}{\bm{\xi}}              
\newcommand{\bOmega}{\bm{\Omega}}         

\newcommand{\norm}[1]{\left\lVert #1 \right\rVert}
\newcommand{\inner}[2]{\langle #1,\, #2 \rangle}
\newcommand{\grad}{\nabla}

\section{Introduction}
\label{sec:introduction}

Federated learning (FL) enables multiple participants to collaboratively train a machine learning model while keeping their training data local~\cite{9464278}. The participants repeatedly train on their local data and exchange model information according to a communication protocol. In conventional centralized federated learning (CFL), this communication structure can be represented as a star graph and clients communicate with a central server, which is the only node performing aggregation. Decentralized federated learning (DFL) generalizes this setting by distributing aggregation across participants, which exchange and combine models according to a general communication topology~\cite{gabrielli2026surveydecentralizedfederatedlearning}.

\begin{figure}[t]
    \centering
    \includegraphics[width=1\linewidth]{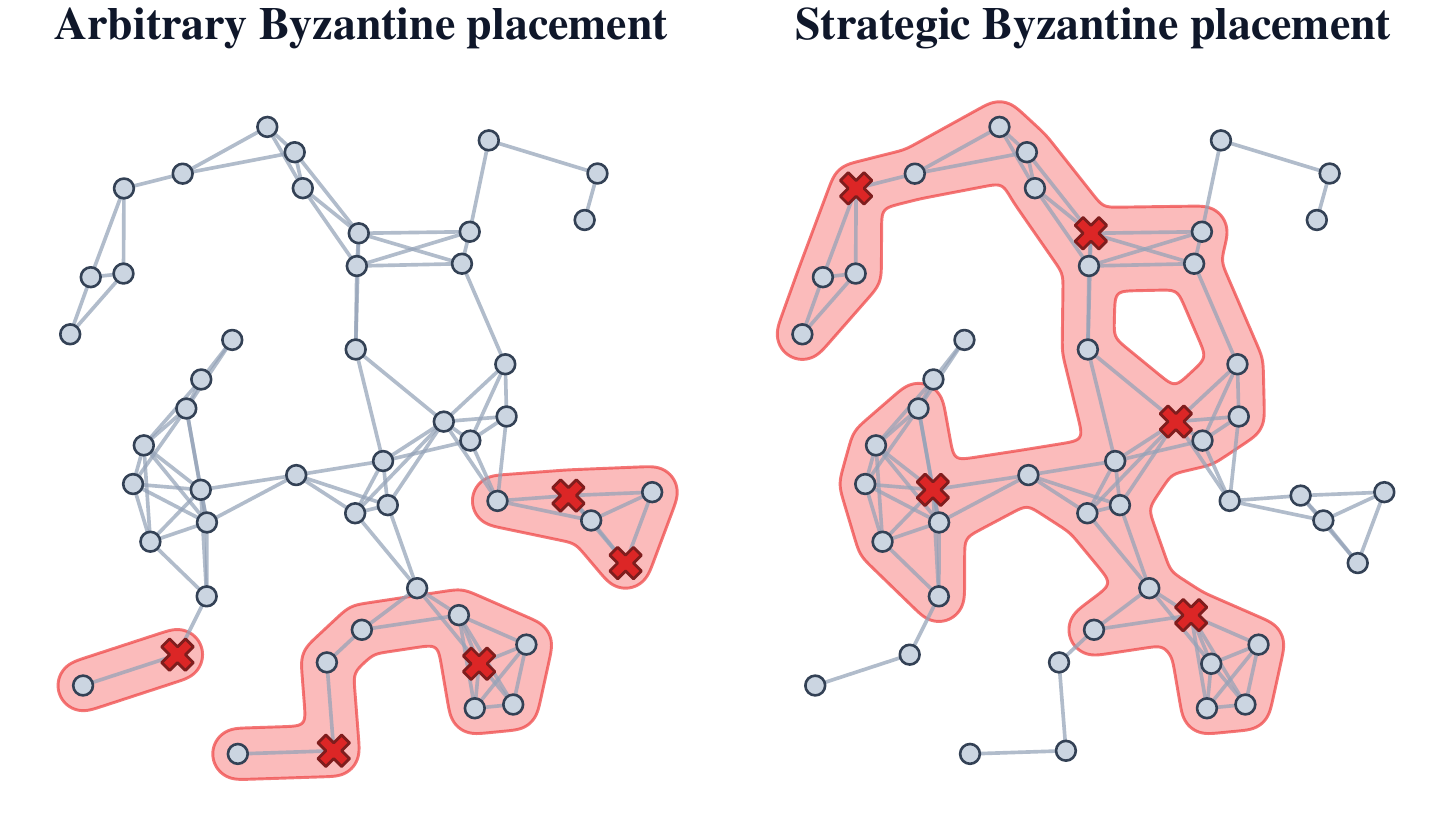}
    \caption{Byzantine placement as an adversarial dimension in decentralized federated learning. The two executions use the same communication topology, attack, and number of Byzantine participants; only the compromised nodes differ. A poorly positioned Byzantine set exposes a limited portion of the network, whereas a strategic placement allows malicious influence to reach substantially more honest participants within the same training horizon.}
    \label{fig:byzanine-placement-cover}
\end{figure}

This generalization introduces an additional adversarial dimension because, while in CFL every client reaches all the others in one aggregation step, in the general DFL topology different participants occupy different positions in the communication graph, and malicious information propagates through successive aggregations before reaching distant nodes. Consequently, two Byzantine sets of the same cardinality may have substantially different effects on the honest network; Figure~\ref{fig:byzanine-placement-cover} illustrates this mechanism.

This raises a basic question: \emph{if an adversary can compromise $m$ participants, which ones should it choose?} Existing Byzantine threat models and evaluations provide no common answer. Byzantine participants are variously selected at random, fixed beforehand, positioned ad hoc, or chosen according to topological structural properties~\cite{guoByzantineResilientDecentralizedStochastic2022,heByzantineRobustDecentralizedLearning2023,fangByzantineRobustDecentralizedFederated2024,bhattacharyaImpactNetworkTopology2024}. Recent work has started to consider strategic node placement~\cite{piasecznyAdversarialNodePlacement2025}, but a general criterion for selecting a Byzantine set according to its actual finite-time influence on decentralized training is still missing.

In this paper, we argue that this is not only an experimental-design detail. If the adversary is able to choose which participants to compromise, evaluating an attack under an arbitrary Byzantine placement may substantially underestimate its actual capability. Conversely, a defense may appear effective because the selected Byzantine nodes are poorly positioned to influence the honest network, rather than because the defense successfully limits malicious updates. Thus, the placement of Byzantine participants should itself be considered part of the threat model.

To address these issues, we formulate Byzantine node placement as an optimization problem. Given a communication topology, a Byzantine budget $m$, and a training horizon $T$, the adversary seeks the set of compromised participants that maximizes its impact on honest nodes. We therefore introduce \emph{Byzantine Placement Influence} (BPI), a topology-based proxy that measures the cumulative finite-time exposure of honest participants to a Byzantine set without observing the actual training dynamics. BPI models how malicious influence propagates over multiple communication hops and evaluates the compromised set jointly. Based on this objective, we develop two efficient placement strategies that identify high-influence Byzantine sets without executing the underlying training task.

Our main contributions are:
\begin{itemize}
    \item We formalize Byzantine node placement as an explicit adversarial
    dimension in DFL, where the attacker selects a set of $m$ compromised
    participants to maximize its finite-time impact on honest nodes.

    \item We introduce Byzantine Placement Influence (BPI), a finite-time,
    set-level measure of Byzantine exposure derived from the gossip
    communication dynamics, together with efficient algorithms for identifying
    high-influence placements.

    \item We show experimentally that BPI-guided placement consistently
    strengthens attacks across heterogeneous communication topologies and
    transfers across different poisoning objectives.

    \item We demonstrate that Byzantine placement can substantially alter the
    apparent robustness of Byzantine-robust aggregation, showing that fixed or
    arbitrary attacker configurations can lead to overly optimistic security
    evaluations.
\end{itemize}

\section{Related Work}
\label{sec:related-works}
The communication topology is a central component of decentralized learning, as it determines which nodes exchange models and how information propagates. Nevertheless, the literature on Byzantine decentralized learning typically treats the position of Byzantine nodes as a secondary experimental detail. Recent works~\cite{gabrielli2026surveydecentralizedfederatedlearning,piasecznyAdversarialNodePlacement2025,gaucherUnifiedBreakdownAnalysis}, however, explicitly raise the question of which nodes an attacker should corrupt in order to maximize the attack's spread over the network. In the following, we review the topology and placement assumptions adopted by works in Byzantine distributed learning that propose attacks or robust aggregation strategies.

\subsection{Topology and Byzantine Placement in Decentralized Learning}
\begin{table*}[htb!]
    \centering
    \begin{tabular}{c|cc}
    \toprule
         Paper & Topology & Byzantine Placement \\
        \midrule
        LEARN~\cite{el-mhamdiCollaborativeLearningJungle2021a} & FC & Irrelevant \\
        UBAR~\cite{guoByzantineResilientDecentralizedStochastic2022} & Random Graph & Random \\
        ClippedGossip~\cite{heByzantineRobustDecentralizedLearning2023} & Dumbbell, SW, Torus, Ring & Random, Ad hoc \\
        Bhattacharya \textit{et al.}~\cite{bhattacharyaImpactNetworkTopology2024} & SW, SF & Random, Ad hoc (SW), Degree (SF) \\
        BALANCE~\cite{fangByzantineRobustDecentralizedFederated2024} & Regular, FC, Erdős--Rényi, SW, Ring & Ad hoc \\
        CS$_+$-RG~\cite{gaucherUnifiedBreakdownAnalysis} & TW, Erdős--Rényi & Ad hoc \\
        SelfishAttack~\cite{jiaCompetitiveAdvantageAttacks2023} & FC, Random Graph & Last $m$ nodes \\
    \bottomrule
    \end{tabular}
    \caption{Topologies and Byzantine placements considered in the literature on Byzantine distributed learning. Acronyms stand for: Fully-Connected (FC), Scale-Free (SF), Small World (SW), and Two Worlds (TW).}
    \label{tab:related-works}
\end{table*}

In fully-connected topologies, Byzantine placement is essentially irrelevant. For instance, LEARN~\cite{el-mhamdiCollaborativeLearningJungle2021a} considers a fully-connected network, where all nodes are structurally equivalent and the attacker does not gain any topological advantage. Similarly, SelfishAttack~\cite{jiaCompetitiveAdvantageAttacks2023} mainly assumes a fully-connected network, with selfish clients modeled as a fixed subset of participants, namely the last $m$ clients in the indexing. SelfishAttack also considers a partially connected random graph, but the selfish clients are not selected according to their topological position.

Other works evaluate Byzantine robustness on sparse or random decentralized networks, but without optimizing the Byzantine set. UBAR~\cite{guoByzantineResilientDecentralizedStochastic2022} uses random graphs generated from a connection probability and randomly adds Byzantine nodes to the benign network. BALANCE~\cite{fangByzantineRobustDecentralizedFederated2024} primarily evaluates regular graphs and additionally considers fully-connected, Erdős--Rényi, small-world, ring, and random graphs with different fractions of malicious--benign edges. However, malicious nodes are fixed in the evaluated graph instances, with no topology-aware selection rule.

Some works use topology-specific or ad-hoc placements. ClippedGossip~\cite{heByzantineRobustDecentralizedLearning2023} evaluates dumbbell graphs, small-world graphs, torus graphs, and rings. In the small-world and torus experiments, Byzantine workers are not selected among the existing honest nodes; instead, the honest topology is first constructed, and additional Byzantine workers are attached to randomly selected regular workers. In other experiments, Byzantine nodes are added in ad-hoc positions to illustrate specific robustness or breakdown phenomena. Notably, its analysis emphasizes local Byzantine influence through the fraction of incident aggregation weight controlled by Byzantine neighbors, rather than only the global number of Byzantine nodes. Bhattacharya \textit{et al.}~\cite{bhattacharyaImpactNetworkTopology2024} study small-world and scale-free networks, placing Byzantine nodes on nodes incident to rewired Watts--Strogatz edges and on high-degree nodes in scale-free graphs; these strategies remain specific to the considered graph families. CS$_+$-RG~\cite{gaucherUnifiedBreakdownAnalysis} analyzes Byzantine-robust gossip on Two Worlds and Erdős--Rényi graphs; in the former, one whole clique is Byzantine, while in the latter each node is adjacent to four malicious nodes.

We summarize these choices in Table~\ref{tab:related-works}. Overall, there is no common placement convention when benchmarking Byzantine attacks and defenses, making results harder to compare and, more importantly, generally leaving the choice of which participants to compromise outside the attacker's optimization problem.

\noindent{\textbf{Strategic Byzantine placement.}}
Two recent works explicitly study Byzantine placement as an adversarial decision. Syros \textit{et al.}~\cite{syrosBackdoorAttacksPeer2025} investigate the effect of compromising structurally influential participants on backdoor propagation in peer-to-peer federated learning. They consider node-wise graph measures, including degree, Effective Network Size, PageRank, and clustering coefficient, and show that the choice of compromised participants can substantially affect backdoor effectiveness. Their placement criteria, however, are static structural heuristics and are studied specifically in the context of backdoor attacks.

Piaseczny \textit{et al.}~\cite{piasecznyAdversarialNodePlacement2025} formulate Byzantine placement as a coordinated adversarial problem and introduce MaxSpAN-FL, a heuristic strategy that favors well-separated attackers. MaxSpAN-FL constructs BFS-based influence regions around candidate nodes and greedily minimizes their overlap, using shortest-path separation as a proxy for adversarial influence. Their evaluation is limited to a single FGSM-based data-poisoning attack and three graph families (directed geometric, Erd\H{o}s--R\'enyi, and preferential-attachment graphs).

Our work builds on this observation but takes a different approach. Instead of using node centrality or shortest-path separation as a proxy for attack influence, BPI derives the placement objective from the gossip dynamics themselves, providing a common finite-time criterion for evaluating a Byzantine set across different communication topologies and poisoning objectives.

\section{Background and Preliminaries}
\label{sec:setup_prelim}

We consider a network composed of $n$ nodes disposed on an undirected graph $\calG = (\calV,\calE)$ where $|\calV| = n$. Each node optimizes a local objective function $f_i: \R^p \rightarrow \R$ and holds a local parameter vector $\btheta \in \R^p$. 

Participants are interested in solving
\begin{equation}
    \btheta^*_{\calH} = \argmin_{\btheta} \left[ \frac{1}{|\calH|} \sum_{i \in \calH} f_i(\btheta) \right]
\end{equation}
where $\calH \subseteq \calV$ denotes the set of honest nodes. 

At each round $t = 0, \dots, T-1$, every node $i$ produces its local model update $\btheta_i^{(t)}$, e.g., by running SGD locally. 
Node $i$ then receives all parameters from its neighbors $\calN(i)$ and linearly combines them:
\begin{equation}
    \btheta^{+}_i = \sum_{j \in \calN(i)\cup\{i\}} w_{ij}\btheta_j.
\end{equation}
Each $w_{ij}$ forms a doubly-stochastic gossip matrix $\bW = [w_{ij}] \in [0,1]^{n \times n}$ and is governed by the following equation:
\begin{equation}
\label{eq:mixing-weights}
w_{ij} =
    \begin{cases}
        \dfrac{1}{1+\max(d_i,d_j)} & (i,j) \in \calE, \\[4pt]
        1 - \sum_{\ell \in \calN(i)} w_{i\ell} & i = j, \\[2pt]
        0 & \text{otherwise}
    \end{cases}
\end{equation}
where $d_i$ denotes the degree of node $i$. By construction $\bW\bone = \bone$ and $\bone^\top \bW = \bone^\top$, and $\lim_{t \to \infty} \bW^t = \tfrac{1}{n}\bone\bone^\top$; meaning that messages eventually spread uniformly across the network, assuming $\calG$ is connected.

\begin{remark}[CFL as a special case]
CFL corresponds to the special case where $\bW$ is uniform over all clients -- i.e., physically a star topology, but functionally equivalent to a fully-connected graph -- since the server places every client at one-hop distance from every other. Under this $\bW$, Byzantine \emph{placement} is irrelevant, because if the server fails to filter an attack, the poisoned update reaches the entire network in a single round regardless of which clients are compromised. 
However, this assumption breaks down as soon as $\bW$ is no longer uniform, i.e., in genuinely decentralized topologies. 
\end{remark}

\subsection{Byzantine Threat Model}
\label{sec:threat}

For the remainder of this work, we consider the scenario in which an external attacker selects a set $\calB \subset \calV$, with $|\calB|=m$ and $\calH \cap \calB = \emptyset$, of Byzantine nodes before training starts. Byzantine nodes deviate from the protocol described above by performing model and data poisoning attacks. 

The attacker operates under the following assumptions:

\begin{description}
    \item[\textbf{(A1) Fixed topology.}] The topology does not change after training starts and cannot be modified by the attacker.
    \item[\textbf{(A2) Topology knowledge.}] The adversary knows the communication topology $\calG$, and every Byzantine node knows the complete set $\calB$ of Byzantine participants.
    \item[\textbf{(A3) Non-robust aggregation.}] Honest nodes do not employ robust aggregation functions, so the mixing matrix $\bW$ remains linear and depends only on the topology.
\end{description}

\noindent{\textbf{Justifying (A1).}}
We consider a static communication graph, consistently with the standard formulation of decentralized gossip learning~\cite{gaucherUnifiedBreakdownAnalysis,heByzantineRobustDecentralizedLearning2023}. This assumption allows us to isolate the effect of the placement of Byzantine nodes. If communication links could change during training or be modified by the adversary, the resulting attack effectiveness would conflate Byzantine placement with topology manipulation, and the propagation dynamics would be governed by a time-varying sequence of mixing matrices rather than by a single matrix $\bW$.

\noindent{\textbf{Justifying (A2).}} Consistent with prior work on Byzantine robustness in DFL~\cite{gaucherUnifiedBreakdownAnalysis, el-mhamdiCollaborativeLearningJungle2021a, heByzantineRobustDecentralizedLearning2023, raynalCanDecentralizedLearning, pasquiniInsecurityPeertoPeerDecentralized2023}, and to avoid relying on security through obscurity, we assume that the adversary has complete knowledge of the communication topology. In practice, the adversary's ability to reconstruct the network depends on the actual implementation and characterizing such discovery mechanisms is orthogonal to our objective; therefore we leave it outside the scope of this work.


\noindent{\textbf{Justifying (A3).}}
We adopt (A3) to isolate the effect of Byzantine placement from the dynamics introduced by robust aggregators and filtering mechanisms~\cite{blanchardMachineLearningAdversaries,mhamdiHiddenVulnerabilityDistributed2018,yinByzantineRobustDistributedLearning2021,caoFLTrustByzantinerobustFederated2022,zhang2022fldetector,gabrielliSecuringFederatedLearning2025}. Under plain averaging, communication is described by the fixed mixing matrix $\bW$, allowing us to separate the topological propagation of Byzantine influence from the particular attack realization. With robust aggregation, instead, whether a received model is accepted, rejected, clipped, or reweighted depends on its relation to the models observed during training. The resulting effective mixing operator is therefore nonlinear and time-varying, and depends on the attack, local data, optimization trajectory, and possibly previous rounds.

Moreover, robust aggregation does not provide a universal elimination of Byzantine influence, since adaptive poisoning attacks can circumvent several distance- and statistics-based defenses~\cite{fangLocalModelPoisoning,baruch2019neurips, shejwalkarManipulatingByzantineOptimizing2021}, while history-aware attacks and recent theoretical results expose broader limitations of robust aggregation strategies~\cite{karimireddyLearningHistoryByzantinea, raynalConflictRobustnessLearning2025}. Modeling a specific robust aggregator inside BPI would consequently make the placement criterion defense- and attack-specific rather than topology-only. We therefore derive BPI under linear aggregation and separately evaluate in Section~\ref{subsec:exp-q4} whether its placement recommendations transfer to Byzantine-robust aggregation.

\noindent{\textbf{Perturbation model.}} To keep the placement problem separate from the choice of poisoning strategy, Byzantine nodes transmit messages of the form $\tilde{\btheta}_i^{(t)} = \btheta_i^{(t)} + \alpha \mathbf{u}$, where $\mathbf{u}$ is a unit direction ($\norm{\mathbf{u}} = 1$) and $\alpha > 0$ is the attack strength. Since we can decompose any poisoning attack into a clean local update plus an adversarial perturbation, and we are concerned only with how that perturbation propagates through the network, we abstract away from how $\alpha$ and $\mathbf{u}$ are generated.

\noindent{\textbf{Byzantine drift.}} We model the effect of an attack on node $i$'s model as $\bdelta_i^{(t)} = \tilde{\btheta}_i^{(t)} - \btheta_i^{(t)}$, the deviation between the model obtained under a placement $\calB$ and the model obtained in a clean run with identical initialization, data partition, and randomness but $\calB = \emptyset$. Stacking all node drifts gives $\bDelta^{(t)} = [\bdelta_1^{(t)}, \dots, \bdelta_n^{(t)}]^\top$, and the empirical honest-node drift of the network is
\begin{equation}
\label{eq:empirical-drift}
    D_{\calH}^{(t)}(\calB) = \frac{1}{|\calH|} \sum_{i \in \calH} \norm{\bdelta_i^{(t)}}.
\end{equation}

\noindent{\textbf{Attacker's objective.}} Putting (A1)--(A3) together, the attacker chooses the placement $\calB$, with $|\calB| = m$ fixed, that maximizes the expected honest-node drift after $T$ rounds:
\begin{equation}
\label{eq:attacker-obj}
    \calB^* = \argmax_{|\calB|=m} \; \E\!\left[D_{\calH}^{(T)}(\calB;\, \xi)\right],
\end{equation}
where $\xi$ collects the exogenous randomness of training (initialization, mini-batch sampling, data partitioning).

Directly solving Eq.~\ref{eq:attacker-obj} is generally computationally intractable. Evaluating the objective for a single candidate placement $\calB$ requires estimating the expected honest-node drift $D_{\calH}^{(T)}(\calB;\,\xi)$, which entails running training twice -- once with $\calB=\emptyset$ and once with the candidate Byzantine placement $\calB$ -- and repeating this process across multiple random seeds to average over the randomness $\xi$.
Moreover, since there are $\binom{n}{m}$ possible placements, exhaustive evaluation and optimization quickly become infeasible even for moderately sized graphs.

To address this challenge, Section~\ref{sec:problem} introduces \textit{Byzantine Placement Influence} (BPI), a tractable proxy objective for identifying influential Byzantine placements.
\section{Byzantine Placement Influence}
\label{sec:problem}

\subsection{From Empirical Drift to a Tractable Proxy}
\label{subsec:bpi-motivation}

\begin{figure}[htb!]
    \centering
    \includegraphics[width=0.5\linewidth]{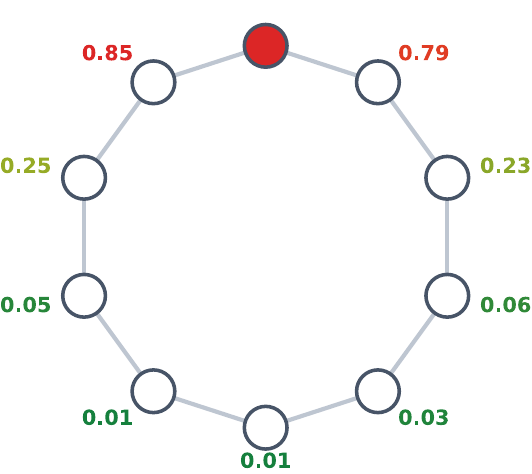}
    \caption{BadNet backdoor propagation on the ring topology. The Byzantine source node is shown in red, while honest nodes are shown in white. Values around honest nodes report final local attack success rate (ASR) on triggered test samples; label color maps lower ASR to green and higher ASR to red.}
    \label{fig:asr-ring-badnets}
\end{figure}

In practical scenarios, DFL training operates over a finite number of rounds $T$, and the effectiveness of a Byzantine placement depends critically on how quickly poisoned mass propagates through the gossip network within that finite time horizon. This is particularly relevant when attacks add only a small bias over the clean baseline, which may not survive the training dynamics if Byzantine sources are located in low-influence regions of the graph. For example, backdoor attacks are sensitive to their positioning in the graph, as their effect survives only for some hops from the source (see Fig.~\ref{fig:asr-ring-badnets}). This is the decentralized analogue of the client-selection problem in CFL, where a client-injected backdoor must survive over several communication rounds~\cite{faraounFLAREFederatedLearning2025,zhangNeurotoxinDurableBackdoors2022,wangAttackTailsYes2020,sunCanYouReally2019,Xie2020DBA:}.

These observations motivate us to define a tractable, topology-aware score that captures finite-time Byzantine exposure to honest nodes without requiring any training runs with the aim of maximizing the adversarial objective.

\subsection{Drift Dynamics and the BPI Score}
\label{subsec:bpi-definition}

Under assumptions (A1)--(A3) and the perturbation model of Section~\ref{sec:threat}, to obtain a topology exposure proxy we consider the linearized gossip component of the drift dynamics:
\begin{equation}
\label{eq:drift-recursion}
    \sbDelta^{(t+1)} = \bW\sbDelta^{(t)} + \alpha\,(\bW\bone_\calB) \otimes \mathbf{u},
\end{equation}
where $\bone_\calB \in \{0,1\}^n$ is the indicator vector of the Byzantine set. The first term propagates the drift accumulated up to round $t$ through one gossip step; the second term injects a fresh perturbation $\alpha\mathbf{u}$ at each Byzantine source, weighted by how much each honest node listens to it.
Specifically, $(\bW\bone_\calB)_i = \sum_{k \in \calB} w_{ik}$ is the total mixing weight that node $i$ assigns to its Byzantine neighbors. Assuming no drift at initialization ($\sbDelta^{(0)} = \mathbf{0}$, since both the clean and poisoned runs share identical initialization under $\calB=\emptyset$ and $\calB\neq\emptyset$ respectively), the recursion unrolls in closed form to:
\begin{equation}
\label{eq:drift-closed-form}
    \sbDelta^{(T)} = \alpha \sum_{t=1}^{T} (\bW^t \bone_\calB) \mathbf{u}.
\end{equation}
Applying the same honest-node drift measure of Eq.~\ref{eq:empirical-drift} to the linearized dynamics, we define the corresponding reference drift as:
\begin{equation}
\label{eq:drift-factored}
    \hat{D}_\calH^{(T)}(\calB) = \frac{\alpha}{|\calH|}\, \bone_\calH^\top \left(\sum_{t=1}^{T} \bW^t\right) \bone_\calB.
\end{equation}
This motivates the following definition.

\begin{definition}[Byzantine Placement Influence (BPI)]
\label{def:bpi}
Given a mixing matrix $\bW$, a Byzantine set $\calB\subseteq\calV$, and a training horizon $T$, the BPI score is computed as:
\begin{equation}
\label{eq:phi_T}
\Phi_T(\bW,\calB)
=
\frac{1}{|\calH|}
\bone_\calH^{\top}
\left(\sum_{t=1}^{T}\bW^{t}\right)
\bone_\calB.
\end{equation}
\end{definition}

The BPI score has a natural interpretation: the $(i,k)$-entry of $\bW^t$ measures the cumulative mixing weight that flows from node $k$ to node $i$ over exactly $t$ gossip steps, and the sum $\sum_{t=1}^T \bW^t$ aggregates this over the entire training horizon. BPI is therefore the average total exposure of honest nodes to Byzantine sources accumulated over $T$ rounds, weighted by the gossip matrix. Note that BPI depends only on $\bW$, $\calB$, and $T$; it requires no training runs and no knowledge of the attack direction $\mathbf{u}$ or local data distributions.

\noindent{\textbf{Comparison with graph-centrality heuristics.}} Standard heuristics for identifying influential nodes -- i.e., degree, betweenness, eigenvector centrality -- score nodes independently of one another and of the training horizon. BPI differs in two key respects. First, it is \emph{set-valued}: BPI evaluates the joint diffusion of the entire Byzantine set, capturing overlap effects that per-node scores miss. For instance, two high-degree nodes adjacent to the same region may expose fewer honest nodes than two lower-degree nodes placed in well-separated parts of the graph. Second, it is \emph{finite-time}: it accounts for how far Byzantine mass actually travels within $T$ rounds.

\subsection{Relaxing the Attacker's Objective with BPI}
\label{subsec:bpi-objective}

From Eq.~\ref{eq:drift-factored}, maximizing $D_\calH^{(T)}(\calB)$ over placements of size $m$ is approximated by maximizing $\Phi_T(\bW,\calB)$. This yields a tractable reformulation of the attacker's objective (Eq.~\ref{eq:attacker-obj}):
\begin{equation}
\label{eq:opt-phi}
    \calB^* \approx \argmax_{\calB \subseteq \calV,\; |\calB|=m} \Phi_T(\bW,\calB).
\end{equation}
The approximation sign reflects two sources of gap relative to Eq.~\ref{eq:attacker-obj}: (i) BPI is a linearization that ignores the training dynamics beyond the gossip step (local SGD noise, data heterogeneity), and (ii) it abstracts away the exogenous randomness $\xi$ by construction. We show in Section~\ref{sec:experiments} that larger BPI values correspond to stronger empirical influence and that BPI-guided placements remain effective across topologies and attack strategies, supporting its use as a placement criterion.

Nevertheless, exact optimization of Eq.~\ref{eq:opt-phi} quickly becomes computationally impractical. As with Eq.~\ref{eq:attacker-obj}, it requires evaluating all $\binom{n}{m}$ possible Byzantine placements, which is feasible only for small graphs. Section~\ref{sec:method} presents two efficient heuristic algorithms for approximating the optimal solution.

\section{BPI-Guided Byzantine Placement}
\label{sec:method}

Section~\ref{sec:problem} established that optimizing Byzantine placement for an attacker can be relaxed to maximizing $\Phi_T(\bW,\calB)$ over all subsets $\calB \subseteq \calV$ of size $m$ (Eq.~\ref{eq:opt-phi}). Since exact optimization is combinatorially intractable for large graphs, we propose two efficient approximation algorithms: a greedy forward selection procedure and a local search refinement built on top of it.

\subsection{Greedy BPI Selection}
\label{subsec:greedy}

The first algorithm builds $\calB$ incrementally. Starting from $\calB = \emptyset$, at each step it adds the candidate node $r \in \calV \setminus \calB$ that yields the largest marginal increase in $\Phi_T(\bW,\calB)$, repeating until $|\calB| = m$. The procedure is described in Algorithm~\ref{alg:greedy-phi-selection}.

\begin{algorithm}[t]
    \caption{\textsc{Greedy-BPI Selection}}\label{alg:greedy-phi-selection}
    \begin{algorithmic}[1]
        \Require Mixing matrix $\bW$; node set $\calV$; budget $m < |\calV|$; horizon $T$.
        \Ensure $\calB \subset \calV$, $|\calB| = m$.
        \Procedure{\textsc{Greedy-BPI}}{$\bW$, $\calV$, $m$, $T$}
            \State $\calB \gets \emptyset$
            \For{$\_$ \textbf{in} $1, \dots, m$}
                \State $s^* \gets -\infty$; $r^* \gets \texttt{null}$
                \ForAll{$r \in \calV \setminus \calB$}
                    \State $s \gets \Phi_T\!\left(\bW,\, \calB \cup \{r\}\right)$
                    \If{$s > s^*$}
                        \State $s^* \gets s$; $r^* \gets r$
                    \EndIf
                \EndFor
                \State $\calB \gets \calB \cup \{r^*\}$
            \EndFor
            \State \Return $\calB$
        \EndProcedure
    \end{algorithmic}
\end{algorithm}

At each step, the greedy procedure selects the node that maximizes the immediate gain in finite-time exposure. However, this selection is locally optimal with respect to the current partial solution, for two reasons. First, the marginal gain of adding a candidate $r$ depends on the Byzantine nodes already selected. Indeed, adding $r$ both creates new exposure from $r$ to the remaining honest nodes and removes $r$ from $\calH$, so any exposure previously directed at $r$ no longer contributes to $\Phi_T(\bW,\calB)$. Second, different Byzantine sources may induce partially overlapping diffusion patterns: an early greedy choice may block a more effective global configuration by steering the set toward a locally concentrated region of the graph. Consequently, greedy selection is not guaranteed to reach the globally optimal $\calB^*$ of size $m$.

\noindent{\textbf{Computational complexity.}} Each evaluation of $\Phi_T(\bW,\calB)$ costs $\calO(T \cdot |\calE|)$ if $\bW^t \bone_\calB$ is computed iteratively (one sparse matrix-vector product per round) rather than materializing $\bW^t$ explicitly. The greedy loop performs $m$ outer iterations, each evaluating at most $n$ candidates, giving an overall cost of $\calO(m \cdot n \cdot T \cdot |\calE|)$.

\subsection{Local Search Refinement via 1-Swap}
\label{subsec:swap}
To address the limitations of greedy selection, we initialize with the greedy solution and refine it through an exhaustive local 1-swap search. At each iteration, the procedure considers all single node exchanges of the form $\tilde{\calB} = \left(\calB \setminus \{b\}\right) \cup \{r\}$ with $b \in \calB$ and $r \in \calV \setminus \calB$, and accepts the swap that yields the largest improvement in $\Phi_T(\bW,\calB)$. The search repeats until no improving exchange exists. The resulting set is locally optimal with respect to all 1-swap moves.
For every $b \in \calB$ and every $r \in \calV\setminus\calB$,
\begin{equation}
    \Phi_T(\bW, \calB) \geq \Phi_T\!\left(\bW,\, (\calB \setminus \{b\}) \cup \{r\}\right).
\end{equation}
Since each accepted swap strictly increases $\Phi_T(\bW,\calB)$ and the number of feasible placements of size $m$ is finite, the procedure terminates in a finite number of iterations. The full algorithm is given in Algorithm~\ref{alg:swap_phi}.

\begin{algorithm}[t]
\caption{\textsc{Swap-BPI Selection}}
\label{alg:swap_phi}
\begin{algorithmic}[1]
\Require Mixing matrix $\bW$; node set $\calV$; budget $m < |\calV|$; horizon $T$.
\Ensure $\calB \subset \calV$, $|\calB| = m$.
\Procedure{\textsc{Swap-BPI}}{$\bW$, $\calV$, $m$, $T$}
    \State $\calB \gets \textsc{Greedy-BPI}(\bW, \calV, m, T)$
    \State $\text{improved} \gets \texttt{true}$
    \While{$\text{improved}$}
        \State $\text{improved} \gets \texttt{false}$
        \State $b^* \gets \texttt{null}$; $r^* \gets \texttt{null}$; $s^* \gets \Phi_T(\bW, \calB)$
        \ForAll{$b \in \calB$}
            \ForAll{$r \in \calV \setminus \calB$}
                \State $\tilde{\calB} \gets (\calB \setminus \{b\}) \cup \{r\}$
                \State $s \gets \Phi_T(\bW,\, \tilde{\calB})$
                \If{$s > s^*$}
                    \State $s^* \gets s$; $b^* \gets b$; $r^* \gets r$
                    \State $\text{improved} \gets \texttt{true}$
                \EndIf
            \EndFor
        \EndFor
        \If{$\text{improved}$}
            \State $\calB \gets (\calB \setminus \{b^*\}) \cup \{r^*\}$
        \EndIf
    \EndWhile
    \State \Return $\calB$
\EndProcedure
\end{algorithmic}
\end{algorithm}

\noindent{\textbf{Computational complexity.}} Each swap iteration evaluates $\Phi_T(\bW,\calB)$ for all $m(n-m)$ candidate exchanges at cost $\calO(T \cdot |\calE|)$ each, giving $\calO(m(n-m) \cdot T \cdot |\calE|)$ per pass. The number of passes until local optimality is bounded by the number of feasible placements.

\section{Experiments}
\label{sec:experiments}

In this section, we address the following research questions (\textbf{RQ}s):
\begin{enumerate}
    \item[\textbf{RQ1:}] \textit{Does finite-time Byzantine exposure reflect the empirical influence of a placement during decentralized training?}
    \item[\textbf{RQ2:}] \textit{Does BPI provide a consistent placement for Byzantine nodes across heterogeneous communication topologies?}
    \item[\textbf{RQ3:}] \textit{Do BPI-guided placements remain effective when the attacker adopts an objective different from the additive untargeted perturbation used to derive BPI?}
    \item[\textbf{RQ4:}] \textit{Does the advantage of BPI-guided placements persist when honest nodes employ a nonlinear Byzantine-robust aggregation?}
\end{enumerate}

\noindent{\textbf{Dataset and model.}}
Unless otherwise stated, we use MNIST~\cite{lecun-mnisthandwrittendigit-2010} and distribute the training samples i.i.d. across participants. Each client trains a two-layer multilayer perceptron consisting of a 128-unit ReLU hidden layer followed by a 10-unit output layer. Local training minimizes the cross-entropy loss using Adam with learning rate $0.01$, batch size $64$, and one local epoch per communication round.

\noindent{\textbf{Setup.}}
Our main experiments use $n=50$ participants, of which $m=5$ ($10\%$) are Byzantine, and run for $T=30$ communication rounds. Honest nodes aggregate neighbor models using Metropolis weights. We evaluate five experiment seeds, $\{33,99,5,42,123\}$.

Byzantine nodes do not aggregate models received from other Byzantine nodes. This removes Byzantine-to-Byzantine self-reinforcement from the experimental setting and isolates the effect of how malicious sources expose honest participants.

We select six topologies described in Table~\ref{tab:topology-config}. We selected the topology parameters so that all graph families yield broadly comparable sparse networks, with average degrees approximately between four and five. This prevents differences in attack effectiveness from being trivially explained by large differences in network density; at the same time, the resulting graphs retain substantially different structures. DC-SBM, Core-Periphery, Random-Geometric, and Scale-Free are regenerated using the experiment seed; disconnected realizations are resampled until a connected graph is obtained. Dragonfly and Ring-of-Cliques are deterministic and therefore identical across seeds.

\begin{table*}[hbt!]
\centering
\caption{Communication topologies used in the main experiments. All graphs contain $n=50$ nodes.}
\label{tab:topology-config}
\begin{tabular}{lp{10.5cm}}
\toprule
\textbf{Topology} & \textbf{Configuration} \\
\midrule
Ring-of-Cliques &
10 cliques of size 5, each forming a $K_5$, connected in a ring by one inter-clique edge per consecutive pair (110 edges)~\cite{liResistanceTwoNodes2022}. \\
Dragonfly &
10 groups of 5 nodes, each forming a $K_5$. The group-level graph is the 4-regular circulant $C_{10}(1,2)$, with one physical inter-group link for each group-level edge and endpoints assigned round-robin (120 edges)~\cite{kimTechnologyDrivenDragonfly2008}. \\
Scale-Free &
Barabási--Albert preferential attachment in which each new node attaches to two existing nodes~\cite{barabasiEmergenceScalingRandom1999}. \\
DC-SBM &
Five communities of 10 nodes, with $p_{\mathrm{in}}=0.4$ and $p_{\mathrm{out}}=0.025$. Node propensities are sampled from a log-normal distribution with $\sigma=0.6$ and normalized to unit mean within each community; edge probabilities are $p_{ij}=\min(1,\theta_i\theta_j p_{\mathrm{in/out}})$~\cite{karrerStochasticBlockmodels2011}. \\
Core-Periphery &
Two-block SBM with 10 core and 40 peripheral nodes, $p_{cc}=0.6$, $p_{cp}=0.15$, and $p_{pp}=0.03$~\cite{borgattiModelsCorePeriphery2000}. \\
Random-Geometric &
Nodes sampled uniformly in $[0,1]^2$ and connected when their Euclidean distance is at most $r=0.17$~\cite{penroseRandomGeometricGraphs2003}. \\
\bottomrule
\end{tabular}
\end{table*}

\noindent{\textbf{Byzantine placement.}}
We compare Greedy-BPI and Swap-BPI with five baselines: uniform random placement, Degree, Betweenness, Eigenvector centrality, and MaxSpAN-FL~\cite{piasecznyAdversarialNodePlacement2025}. Centrality-based strategies select the $m$ nodes with the largest respective centrality score.

For Random, one Byzantine set of $m$ distinct nodes is sampled uniformly without replacement for each topology and experiment seed. Hence, each table entry for Random contains five independently generated placements, one for each seed.

\noindent{\textbf{Poisoning attacks.}}
For the untargeted evaluation, Byzantine nodes inject an aligned additive perturbation. At the beginning of each run, we sample a direction $\mathbf u$ with entries $u_k\sim\mathcal N(0,1)$ using the experiment seed and normalize it such that $\|\mathbf u\|_2=1$. The same direction is shared by all Byzantine nodes.

When Byzantine node $i$ attacks for the first time, we record its model norm $\|\boldsymbol{\theta}_{i,1}\|_2$ and freeze this value for the remainder of the run. At every subsequent communication round, the node publishes
\[
    \widetilde{\boldsymbol{\theta}}_{i,t}
    =
    \boldsymbol{\theta}_{i,t}
    +
    \alpha \|\boldsymbol{\theta}_{i,1}\|_2 \mathbf u,
\]
where $\alpha=64$ in the main experiments. Thus, all malicious perturbations are directionally aligned while their magnitude is normalized to the initial model scale of the corresponding attacker. We refer to this attack as FixedBias.

For the targeted evaluation, we employ a classical backdoor attack called BadNets~\cite{guBadNetsIdentifyingVulnerabilities}. Each Byzantine participant poisons a fixed $20\%$ subset of its local training data. Poisoned samples are stamped with a $3\times3$ white square in the bottom-right corner and relabeled to target class $y^*=0$. Malicious clients otherwise follow the same local optimization configuration as honest clients, i.e., one local epoch of Adam training with learning rate $0.01$ and batch size $64$ per communication round.

\noindent{\textbf{Evaluation metrics.}}
For RQ1, we use the empirical honest-node drift $D_{\calH}^{(t)}$ from Eq.~\ref{eq:empirical-drift} and examine whether placements with progressively larger $\Phi_T(\bW,\calB)$ induce correspondingly larger drift trajectories. For untargeted attacks, we measure the honest-node accuracy drop at round $t$ as
\begin{equation*}
    \Delta A_t = \frac{1}{|\calH|} \sum_{i \in \calH} (A_{i,t} - \tilde{A}_{i,t}),
\end{equation*}
where $A_{i,t}$ is node $i$'s clean accuracy and $\tilde{A}_{i,t}$ its accuracy under attack. The main RQ2 comparison reports the final-round value $\Delta A_T$ with $T=30$; the long-horizon FixedBias experiment in RQ4 reports $\Delta A_t$ over training.

For backdoor attacks, we measure the honest-node attack success rate (ASR) on triggered test samples:
\begin{equation}
    ASR_{\mathcal H}^{(t)}
    =
    \frac{1}{|\mathcal H|}
    \sum_{i\in\mathcal H}
    \frac{
    \sum_{(x,y)\in D_{\mathrm{test}}:y\neq y^*}
    \mathbbm{1}\!\left[
    f_i^{(t)}(\tau(x))=y^*
    \right]
    }{
    |\{(x,y)\in D_{\mathrm{test}}:y\neq y^*\}|}.
\end{equation}
Here, $f_i^{(t)}(\cdot)$ is the classifier held by node $i$ at round $t$, $D_{\mathrm{test}}$ is the clean test set, $\tau(x)$ adds the backdoor trigger, and $y^*$ is the attacker's target label.

\subsection{Empirical Validation of Finite-Time Exposure (RQ1)}
\label{subsec:exp-q1}
\begin{figure}[htb!]
    \centering
    \includegraphics[width=1\linewidth]{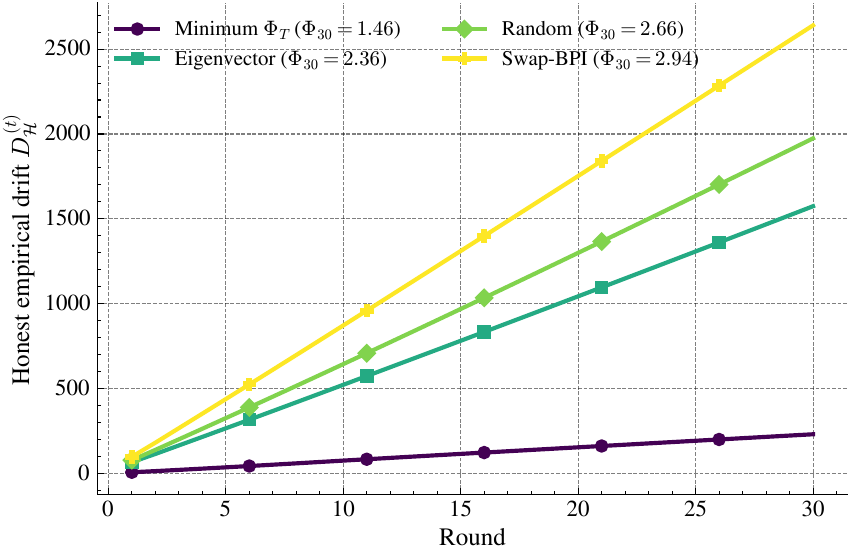}
    \caption{Evolution of the empirical drift $D_{\mathcal H}^{(t)}$ across communication rounds for placements spanning increasing values of $\Phi_T(\bW,\calB)$. Greater exposure results in faster and larger drift of honest models.}
    \label{fig:phi-drift-curves}
\end{figure}

We first investigate whether the ordering induced by BPI is reflected in the actual influence of Byzantine perturbations during training. Figure~\ref{fig:phi-drift-curves} compares four placements spanning progressively larger values of $\Phi_T(\bW,\calB)$ on a 50-node Ring-of-Cliques topology under FixedBias.

The empirical trajectories exhibit the same ordering as BPI. The minimum-exposure placement induces only limited honest-model drift, while Eigenvector, Random, and Swap-BPI produce progressively faster and larger drift accumulation as their finite-time exposure increases. Swap-BPI, which has the largest $\Phi_T(\bW,\calB)$, also produces the largest empirical drift throughout training.

\subsection{Cross-Topology Consistency of BPI (RQ2)}
\label{subsec:exp-q2}

\subsubsection{BPI Across Placement Strategies}
We next study whether a placement strategy that performs well on one graph family remains effective on structurally different networks.

\begin{figure}[htb!]
    \centering
    \includegraphics[width=1\linewidth]{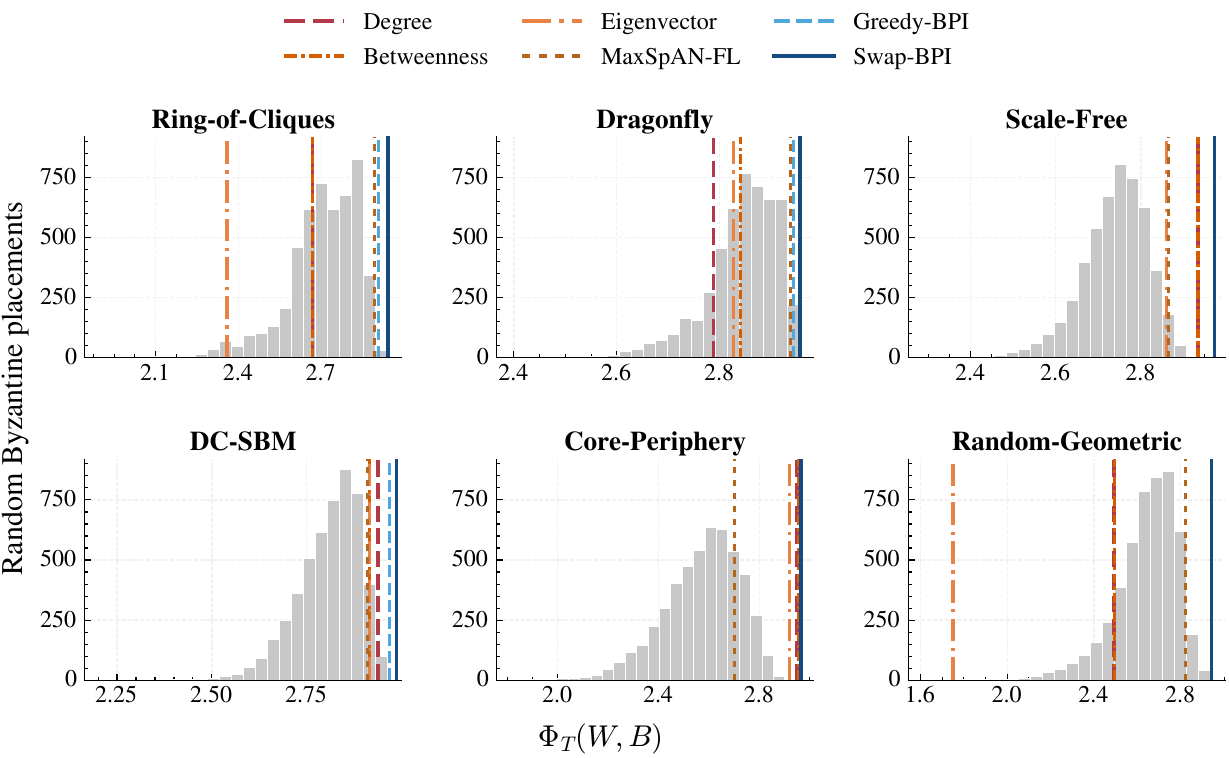}
    \caption{Histograms show \(\Phi_T(W,B)\) for 5000 uniformly sampled placements of five Byzantine nodes on each topology; vertical lines indicate deterministic placement strategies. Lines farther to the right correspond to greater finite-time exposure. We use $n=50$ and \(T=30\).}
    \label{fig:phi-random-histo}
\end{figure}

Figure~\ref{fig:phi-random-histo} first isolates the topology-level objective by comparing $\Phi_T$ obtained by each deterministic placement strategy with the distribution induced by 5000 uniformly sampled Byzantine sets. The relative quality of conventional structural heuristics changes markedly across graph families. A criterion that produces a high-exposure placement on one topology may fall considerably closer to the random-placement distribution on another.

BPI exhibits a qualitatively different behavior. Greedy-BPI and, in particular, Swap-BPI consistently locate placements at the extreme high-exposure end of the distribution across all six graph families.

\subsubsection{Accuracy Drop on Untargeted Attacks}
Table~\ref{tab:fixedbias-placement-t30} shows that no conventional structural heuristic provides a uniformly strong placement across graph families.

Degree and Betweenness are particularly effective on topologies containing pronounced degree heterogeneity or structurally privileged regions. On Scale-Free, they obtain $19.6$ and $19.4$ percentage points of accuracy degradation, respectively, compared with only $14.1$ for Random. A similar pattern appears on DC-SBM and Core-Periphery, where both heuristics approach the BPI-guided placements. This is expected: in these graphs, hubs, high-connectivity nodes, and inter-community positions are informative proxies for malicious influence.

However, the same heuristics transfer poorly to other structures. On Random-Geometric, Degree falls to $15.3$ and Betweenness to $14.0$, compared with $20.8$ for Greedy-BPI. They also provide little advantage over Random on Ring-of-Cliques and Dragonfly.

MaxSpAN-FL exhibits an almost complementary behavior. It performs strongly on Ring-of-Cliques ($20.9$), Dragonfly ($20.7$), and Random-Geometric ($18.4$), but is substantially weaker on Scale-Free ($17.1$), DC-SBM ($16.5$), and Core-Periphery ($17.1$). Eigenvector centrality is less reliable overall and is the weakest strategy on Ring-of-Cliques, Dragonfly, and Random-Geometric.

In contrast, BPI does not rely on any one of these structural signatures. Either Greedy-BPI or Swap-BPI achieves the largest mean degradation on every evaluated topology. This is the main cross-topology result: different structural heuristics become appropriate on different graph families, whereas the propagation-based BPI objective remains consistently effective.

Interestingly, the difference between Greedy-BPI and Swap-BPI is small. Greedy-BPI already attains the strongest observed mean on Dragonfly and Random-Geometric and ties Swap-BPI on DC-SBM, while Swap-BPI is marginally stronger on Ring-of-Cliques, Scale-Free, and Core-Periphery. Their topology-averaged degradations are $20.8$ and $20.9$ percentage points, respectively.

Recall that Swap-BPI is guaranteed to improve or preserve the BPI objective, not the downstream accuracy degradation. Small inversions between Greedy-BPI and Swap-BPI are therefore expected because BPI is a proxy for the complete learning dynamics. The small empirical gap also indicates that most of the benefit comes from optimizing the BPI objective itself.

\begin{table*}[t]
\centering
\caption{Final honest-node accuracy drop $\Delta A_{30}$ under FixedBias for different Byzantine placement strategies. Values are the mean $\pm$ 95\% Student-$t$ confidence interval over five seeds, in percentage points. Higher values indicate a stronger attack. The \textbf{highest} mean in each row is bold; the \underline{lowest} is underlined.}
\label{tab:fixedbias-placement-t30}
\resizebox{\textwidth}{!}{%
\begin{tabular}{lccccc|cc}
\toprule
\textbf{Topology} & \textbf{Random} & \textbf{Degree} & \textbf{Betweenness} & \textbf{Eigenvector} & \textbf{MaxSpAN-FL} & \textbf{Greedy-BPI} & \textbf{Swap-BPI} \\
\midrule
Ring-of-Cliques & $18.2 \pm 1.0$ & $18.4 \pm 2.4$ & $18.4 \pm 2.4$ & $\underline{14.3 \pm 4.5}$ & $20.9 \pm 1.1$ & $21.0 \pm 1.0$ & $\bm{21.2 \pm 1.1}$ \\
Dragonfly & $19.8 \pm 1.3$ & $19.9 \pm 1.6$ & $18.8 \pm 1.2$ & $\underline{18.5 \pm 2.3}$ & $20.7 \pm 1.2$ & $\bm{20.9 \pm 1.1}$ & $20.8 \pm 1.0$ \\
Scale-Free & $\underline{14.1 \pm 3.0}$ & $19.6 \pm 1.0$ & $19.4 \pm 0.9$ & $18.4 \pm 1.1$ & $17.1 \pm 1.3$ & $20.2 \pm 1.0$ & $\bm{20.5 \pm 0.7}$ \\
Degree-Corrected-SBM & $\underline{16.1 \pm 0.5}$ & $20.6 \pm 1.8$ & $20.4 \pm 1.5$ & $19.7 \pm 2.4$ & $16.5 \pm 3.8$ & $\bm{21.3 \pm 1.5}$ & $\bm{21.3 \pm 1.6}$ \\
Core-Periphery & $\underline{13.1 \pm 3.7}$ & $19.3 \pm 0.8$ & $19.8 \pm 0.9$ & $18.8 \pm 0.9$ & $17.1 \pm 2.4$ & $20.6 \pm 0.9$ & $\bm{20.8 \pm 0.9}$ \\
Random-Geometric & $15.2 \pm 1.5$ & $15.3 \pm 1.9$ & $14.0 \pm 2.7$ & $\underline{9.1 \pm 4.5}$ & $18.4 \pm 2.4$ & $\bm{20.8 \pm 0.6}$ & $20.6 \pm 0.8$ \\
\bottomrule
\end{tabular}%
}
\end{table*}

\subsection{Transferability Across Attack Objectives (RQ3)}
The cross-topology instability of the structural baselines becomes even more pronounced under BadNets (Table~\ref{tab:badnets-placement-t30}).

\begin{table*}[t]
\centering
\caption{$ASR_{\calH}$ under BadNets for different Byzantine placement strategies. Entries report the mean $\pm$ 95\% Student-$t$ confidence interval across five seeds. Higher values indicate a stronger attack. The \textbf{highest} mean in each row is bold; the \underline{lowest} is underlined.}
\label{tab:badnets-placement-t30}
\resizebox{\textwidth}{!}{%
\begin{tabular}{lccccc|cc}
\toprule
\textbf{Topology} & \textbf{Random} & \textbf{Degree} & \textbf{Betweenness} & \textbf{Eigenvector} & \textbf{MaxSpAN-FL} & \textbf{Greedy-BPI} & \textbf{Swap-BPI} \\
\midrule
Ring-of-Cliques & $38.6 \pm 6.8$ & $37.0 \pm 6.6$ & $37.0 \pm 6.6$ & $\underline{26.9 \pm 12.6}$ & $42.7 \pm 0.9$ & $\bm{44.6 \pm 1.6}$ & $44.0 \pm 2.5$ \\
Dragonfly & $41.4 \pm 1.8$ & $39.1 \pm 7.4$ & $\underline{31.6 \pm 1.8}$ & $32.6 \pm 9.5$ & $\bm{46.0 \pm 2.0}$ & $44.2 \pm 2.2$ & $45.3 \pm 4.4$ \\
Scale-Free & $\underline{19.7 \pm 8.1}$ & $49.6 \pm 2.1$ & $47.7 \pm 2.1$ & $39.2 \pm 5.4$ & $34.3 \pm 7.4$ & $56.1 \pm 9.5$ & $\bm{59.5 \pm 10.4}$ \\
Degree-Corrected-SBM & $\underline{27.2 \pm 10.2}$ & $56.0 \pm 13.4$ & $53.4 \pm 10.1$ & $47.1 \pm 15.7$ & $29.7 \pm 22.8$ & $63.1 \pm 8.1$ & $\bm{64.7 \pm 8.8}$ \\
Core-Periphery & $\underline{17.0 \pm 9.8}$ & $50.4 \pm 9.0$ & $54.2 \pm 7.5$ & $46.6 \pm 8.2$ & $31.9 \pm 11.9$ & $60.3 \pm 4.8$ & $\bm{60.6 \pm 5.7}$ \\
Random-Geometric & $32.3 \pm 7.8$ & $29.3 \pm 6.8$ & $19.6 \pm 8.4$ & $\underline{14.7 \pm 9.2}$ & $35.6 \pm 8.6$ & $\bm{49.5 \pm 5.5}$ & $48.9 \pm 6.9$ \\
\bottomrule
\end{tabular}%
}
\end{table*}

Degree is highly effective on Scale-Free ($49.6\%$), DC-SBM ($56.0\%$), and Core-Periphery ($50.4\%$), but its effectiveness drops to $29.3\%$ on Random-Geometric. Betweenness follows a similar pattern: it is competitive on Core-Periphery ($54.2\%$) and DC-SBM ($53.4\%$), but reaches only $31.6\%$ on Dragonfly and $19.6\%$ on Random-Geometric.

MaxSpAN-FL again exhibits a different topology preference. It is the strongest strategy on Dragonfly ($46.0\%$) and performs well on Ring-of-Cliques and Random-Geometric, but drops to $34.3\%$ on Scale-Free, $29.7\%$ on DC-SBM, and $31.9\%$ on Core-Periphery. Eigenvector centrality never provides the strongest placement and is particularly weak on Random-Geometric, where it reaches only $14.7\%$ ASR.

Random placement follows the same topology-dependent pattern observed for FixedBias. It remains relatively competitive on Ring-of-Cliques and Dragonfly, but performs very poorly on Scale-Free ($19.7\%$), DC-SBM ($27.2\%$), and especially Core-Periphery ($17.0\%$).

BPI is substantially more stable. A BPI-guided strategy achieves the largest mean ASR on five of the six graph families. On the only exception, Dragonfly, MaxSpAN-FL reaches $46.0\%$ while Swap-BPI reaches $45.3\%$, a difference of only $0.7$ percentage points. Conversely, whenever one of the structural baselines fails to match the graph structure, its degradation can be severe.

This result is particularly significant because BadNets is not described by the aligned additive perturbation model used to derive BPI. The persistence of the same cross-topology behavior therefore indicates that BPI is capturing a property of the communication process itself rather than a peculiarity of FixedBias.

\begin{figure*}[htb!]
    \includegraphics[width=1\textwidth]{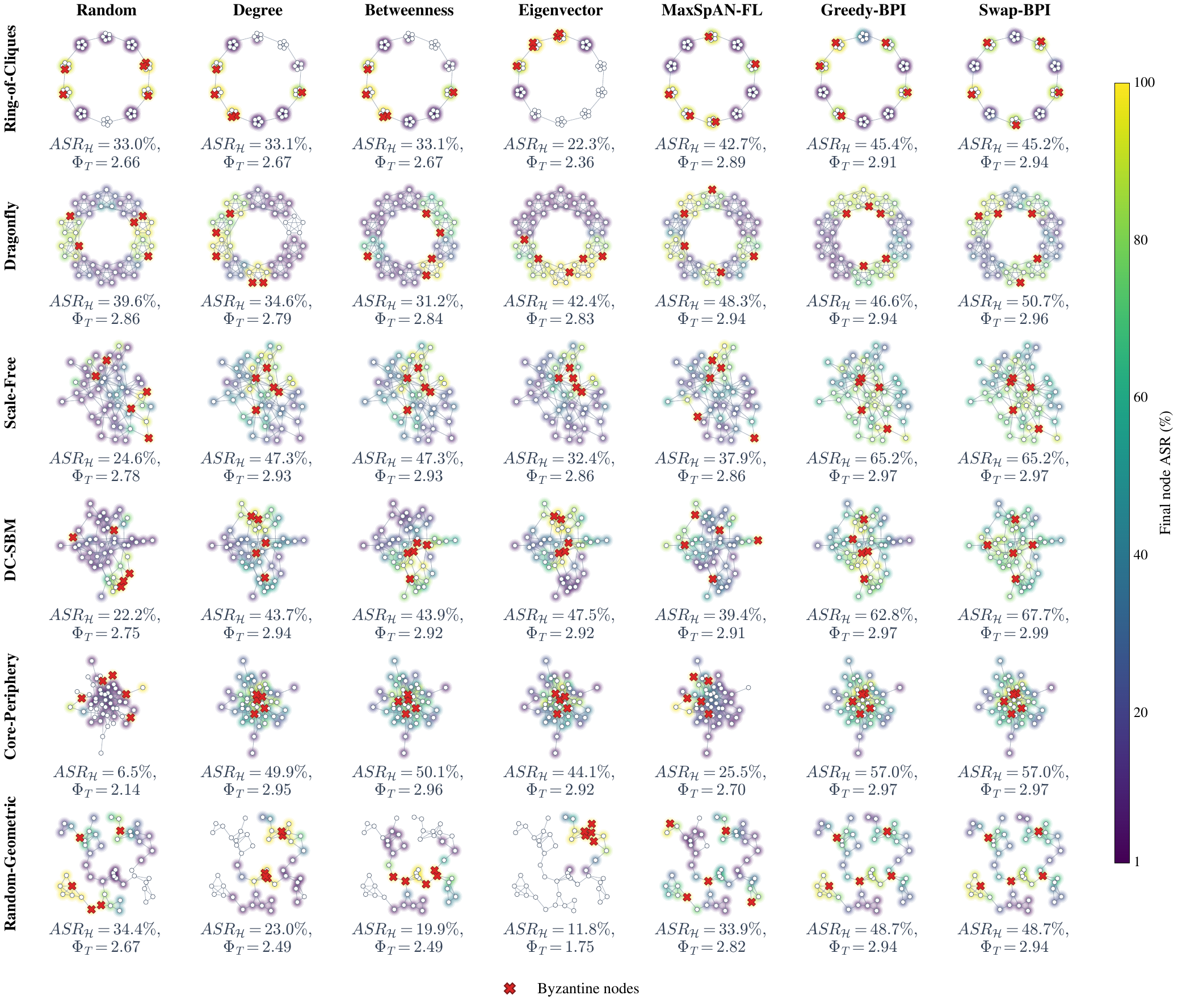}
    \caption{Final node-level ASR across network topologies and Byzantine node placement strategies. Node halos encode the final ASR at round $T=30$; nodes with ASR below $1\%$ are left uncoloured. Each panel reports the mean ASR over honest nodes, $ASR_{\mathcal H}$, and the corresponding $\Phi_T(\bW,\calB)$ score.}
    \label{fig:badnets-node-asr}
\end{figure*}

Figure~\ref{fig:badnets-node-asr} provides a node-level view of this propagation process. Although only Byzantine nodes directly inject poisoned training samples, the learned backdoor is not confined to their immediate neighborhood. Honest participants that aggregate a malicious model partially inherit the backdoor and subsequently propagate it through their own outgoing models. As a result, non-neighboring honest nodes can acquire substantial ASR after several communication rounds.

This behavior illustrates why Byzantine placement cannot be characterized solely through one-hop quantities such as degree. The effect of compromising a node depends not only on how many honest neighbors it directly reaches, but also on how its influence propagates through sequences of honest relays over the finite training horizon. These multi-hop paths are precisely the contribution captured by the powers $\bW^t$ in BPI.

\subsection{Beyond Linear Gossip (RQ4)}
\label{subsec:exp-q4}
Our previous experiments use Metropolis aggregation, matching the linear communication model from which BPI is derived. We finally test whether BPI-guided placement remains effective when the actual aggregation rule is nonlinear.

We consider the Ring and Erdős--Rényi graph configurations and Byzantine placements used in the BALANCE evaluation and compare those placements with Swap-BPI. These experiments contain $n=20$ nodes and $m=4$ Byzantine participants ($20\%$).

We use BALANCE~\cite{fangByzantineRobustDecentralizedFederated2024} with $\alpha_{\mathrm{BAL}}=0.25$, $\kappa=0.5$, $\lambda(t)=t/T$, and $\gamma=1.5$. A neighbor $j$ is accepted by node $i$ whenever
\[
\left\|\btheta_i^{t+1/2}-\btheta_j^{t+1/2}\right\|_2
\le
1.5e^{-0.5t/T}\left\|\btheta_i^{t+1/2}\right\|_2.
\]
The mean of the accepted neighboring models is then combined with the local intermediate model with weights $0.75$ and $0.25$, respectively.

\begin{figure}[htb!]
    \includegraphics[width=1\linewidth]{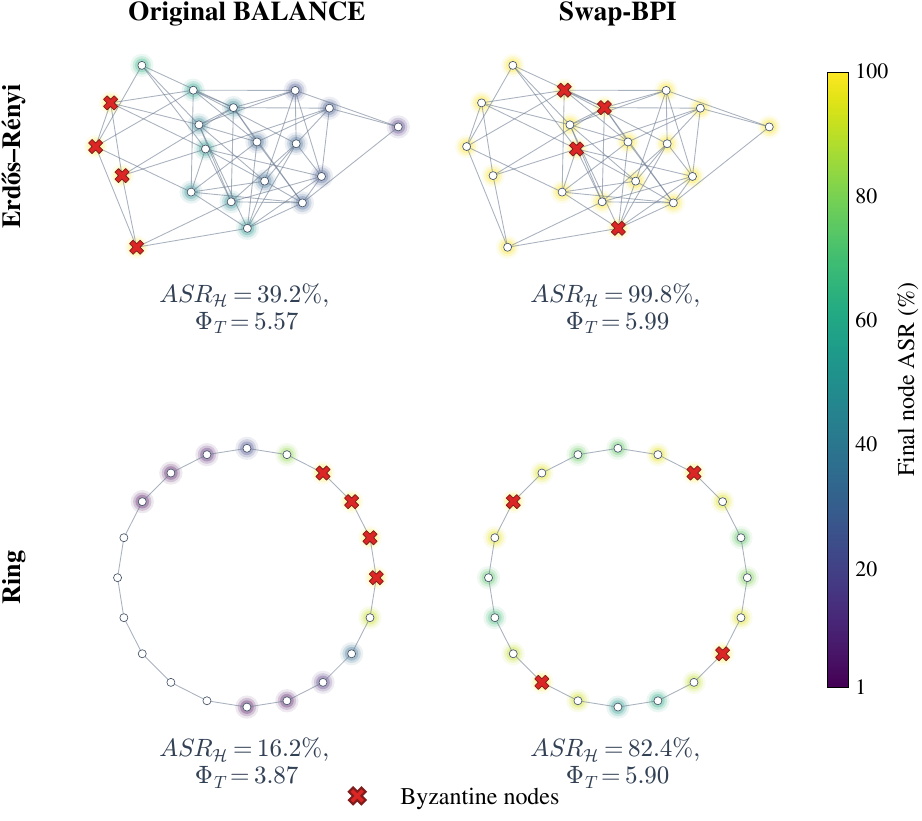}
    \caption{Both the Erdős--Rényi and Ring graphs, together with the Byzantine positioning in the first column, reproduce the topologies and attacker placement used in the original BALANCE paper. Columns compare this positioning with Swap-BPI. Colored halos report final local ASR, while red crosses identify Byzantine nodes. Each panel reports the mean ASR over honest nodes and the corresponding $\Phi_T(\bW,\calB)$ score.}
    \label{img:balance-asr-contour}
\end{figure}

Figure~\ref{img:balance-asr-contour} reveals a substantial placement-induced vulnerability. Under BadNets, the placement used in the original BALANCE evaluation yields a mean honest-node ASR of $39.2\%$ on Erdős--Rényi and $16.0\%$ on the Ring. Changing only the identities of the Byzantine participants to the Swap-BPI placement raises these values to $99.8\%$ and $82.4\%$, respectively.

Thus, the observed robustness of BALANCE is strongly conditional on where those Byzantine nodes are located. The original placement does not expose this worst-case behavior, whereas topology-aware placement reveals a substantially more damaging attack configuration.

Surprisingly, robust aggregation does not necessarily attenuate the topology-induced propagation of the backdoor. Under BadNets, BALANCE can yield higher $ASR_{\calH}$ than plain averaging. This indicates that filtering neighboring models may alter their relative influence in a way that favors malicious updates that remain within the acceptance threshold.

\begin{figure}[htb!]
    \includegraphics[width=1\linewidth]{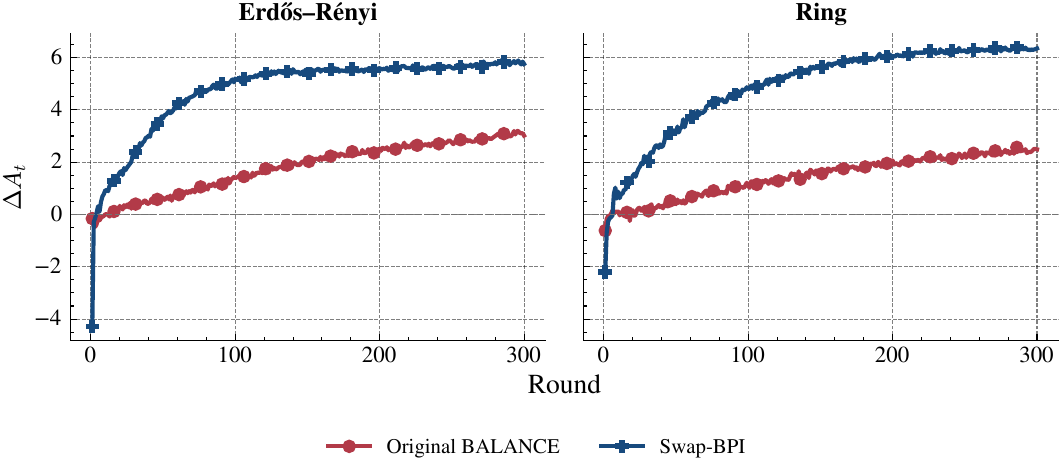}
    \caption{Mean $\Delta A_t$ among honest nodes under BALANCE aggregation for the Byzantine placement used in the original BALANCE paper and Swap-BPI on the Erdős--Rényi and Ring topologies.}
    \label{img:balance-fixedbias}
\end{figure}

We additionally evaluate FixedBias with scale factor $\alpha=2.5$ for $T=300$ rounds. Figure~\ref{img:balance-fixedbias} compares the original BALANCE placement with Swap-BPI. While the original placement produces limited degradation, the BPI-guided placement causes a progressively larger accuracy gap over training, showing that the placement effect also persists for an untargeted attack under nonlinear aggregation.

\section{Conclusion}
\label{sec:conclusion}

This work studies Byzantine node placement as an explicit adversarial dimension in DFL. We formalize the attacker's placement objective and introduce Byzantine Placement Influence (BPI), a finite-time, set-level measure of how strongly a Byzantine set can expose honest participants. BPI separates the structural contribution of the communication topology from the details of a particular poisoning strategy, and the resulting placement algorithms provide a practical way to identify high-influence Byzantine sets without executing training for every candidate placement. Our experiments show that placement can substantially change attack effectiveness across heterogeneous graph families, poisoning objectives, and even under nonlinear Byzantine-robust aggregation. These findings indicate that Byzantine placement should be specified and optimized as part of DFL security evaluation.

Our analysis focuses on static, known communication topologies and derives BPI from nominal linear gossip dynamics. Extending the placement problem to time-varying or partially known topologies, and defense-specific propagation dynamics remains an important direction for future work. Additional deployment constraints can also be incorporated by restricting the feasible placement set; for example, the proposed optimization algorithms can be extended to enforce a bounded local Byzantine population around honest nodes.

\section*{Acknowledgment}
We thank Anthony Di Pietro for contributions to the experimental codebase, including implementations of attack, method, and aggregation components, debugging assistance, and preliminary experiments that informed our investigation of the echo effect.


\bibliographystyle{IEEEtran}
\bibliography{references, refs2}

\end{document}